\documentclass[10pt,journal,letterpaper,compsoc]{IEEEtran}
\ifCLASSOPTIONcompsoc
\else
\fi

\ifCLASSINFOpdf
\else
\fi

\usepackage{graphicx}
\usepackage{color}
\usepackage{amsmath,bm,bbm,dsfont}
\usepackage{amsfonts}
\usepackage{subfigure}
\usepackage{caption}
\usepackage{slashbox}
\usepackage{booktabs}
\graphicspath{{figures/}}

\newenvironment{myitemize}[1][]{
\begin{list}{{#1}} 
    {
     \setlength{\leftmargin}{0cm}      %
     \setlength{\parsep}{0ex}          %
     \setlength{\topsep}{0bp}          %
     \setlength{\itemsep}{0ex}         %
     \setlength{\labelsep}{0.2em}      %
     \setlength{\itemindent}{2em}      %
     \setlength{\listparindent}{1.5em} %
    }}
{\end{list}}

\begin{document}
%
\title{Photo2Relief: Let Human in the Photograph Stand Out}

\author{Zhongping Ji$ ^*$, Feifei Che, Hanshuo Liu, Ziyi Zhao, Yu-Wei Zhang and Wenping Wang  
\IEEEcompsocitemizethanks{\IEEEcompsocthanksitem Zhongping Ji, Feifei Che, Hanshuo Liu and Ziyi Zhao are with the School of Computer Science, Hangzhou Dianzi University, Hangzhou, China. \protect
\IEEEcompsocthanksitem Yu-Wei Zhang is with the School of Mechanical and Automotive Engineering, Qilu University of Technology, Jinan, China. \protect
\IEEEcompsocthanksitem Wenping Wang is with the Department of Computer Science, The University of Hong Kong, Hong Kong, China. \protect
}
\thanks{Corresponding author: Zhongping Ji}}

\markboth{}%
{}
%

\makeatletter
\long\def\@IEEEtitleabstractindextextbox#1{\parbox{0.922\textwidth}{#1}}
\makeatother
\IEEEcompsoctitleabstractindextext{%
\begin{abstract}

In this paper, we propose a technique for making humans in photographs protrude like reliefs. 
Unlike previous methods which mostly focus on the face and head, our method aims to generate art works that describe the whole body activity of the character. 
One challenge is that there is no ground-truth for supervised deep learning. 
We introduce a sigmoid variant function to manipulate gradients tactfully and train our neural networks by equipping with a loss function defined in gradient domain. 
The second challenge is that actual photographs often across different light conditions. 
We used image-based rendering technique to address this challenge and acquire rendering images and depth data under different lighting conditions.
To make a clear division of labor in network modules, a two-scale architecture is proposed to create high-quality relief from a single photograph. 
Extensive experimental results on a variety of scenes show that our method is a highly effective solution for generating digital 2.5D artwork from photographs. 

\end{abstract}

}

\maketitle


\IEEEdisplaynotcompsoctitleabstractindextext

\IEEEpeerreviewmaketitle

\section{Introduction}

\begin{figure}
  \centering
  \subfigure{
    \label{fig:ancientreliefs:a}
    \includegraphics[angle=0,width=1.2085in]{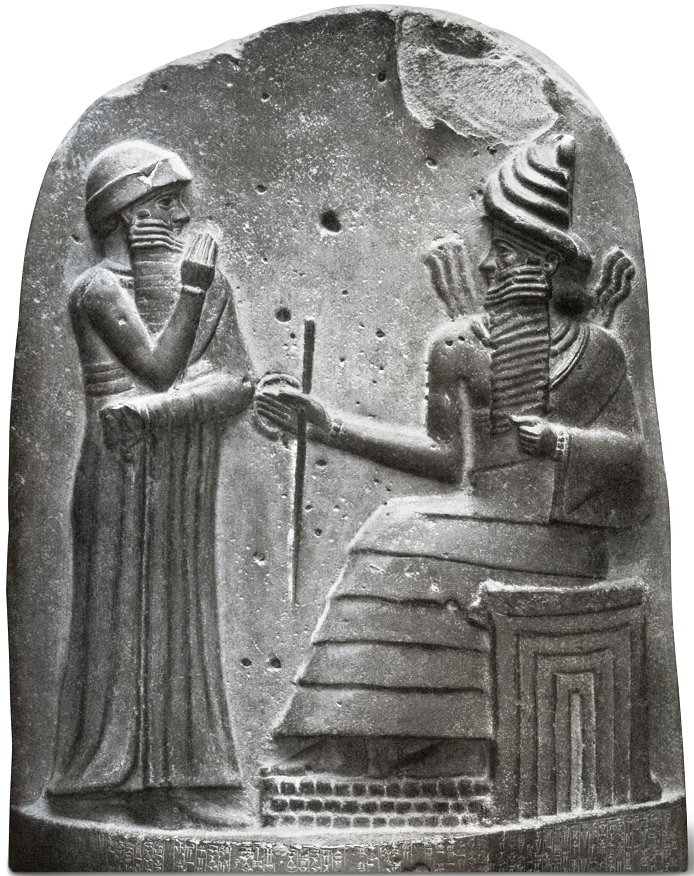}}
    \hspace{-0.01in}
  \subfigure{
    \label{fig:ancientreliefs:b}
    \includegraphics[angle=0,width=2.0in]{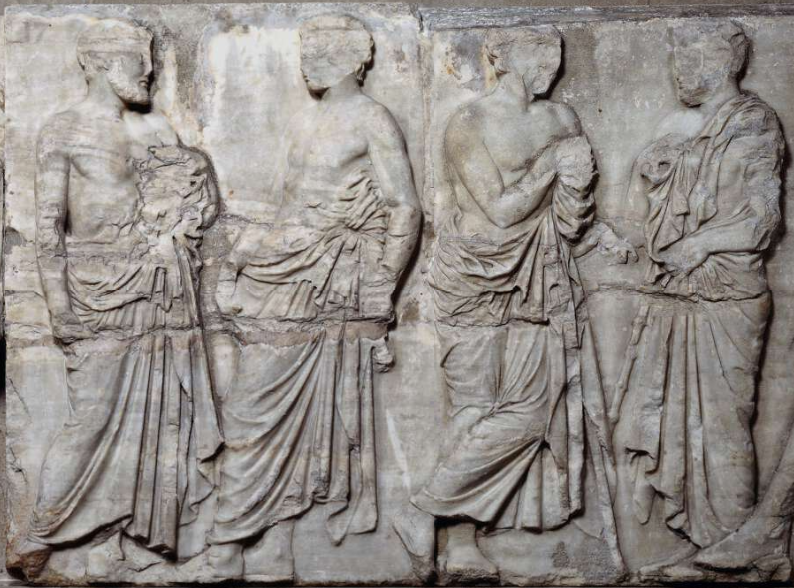}}
  \caption{\label{fig:ancientreliefs}
Reliefs in ancient Babylon and Greek sculptures.
}
\end{figure}

\IEEEPARstart{R}{elief}, as a special art form between drawing and sculpture, has historically been utilized to record diverse human activities. 
Thousands of years ago, relief sculptures were added to the surfaces of stone buildings constructed by ancient Egyptians and Assyrians. 
People carved figures into stone to celebrate the lives of important events and figures. 
Two ancient reliefs carved in stones are shown in Fig. \ref{fig:ancientreliefs}. 
The Babylonian relief occupies the upper part of the Stele of Hammurabi which was built about 3800 years ago. 
Hammurabi is portrayed receiving his royal insignia, the rod and ring, directly from Shamash, the Babylonian god of justice. 
Characters in this art form are more solemn and more impressive than those in general painting. 
Reliefs can also be found in ancient Greek and Roman sculpture, a famous example is the Parthenon frieze featuring relief sculptures.
The example on the right shows a marble relief from the East frieze of the Parthenon. 
The four men draped in simple cloaks were leaning on sticks. 
Although the relief sculptures were destroyed during the war, the outlines and garment wrinkles of the characters were still clearly visible.

For most of the time, the reliefs are produced by skilled artists (such sculptors and engravers), which is a time-consuming task. 
Furthermore, the reliefs produced are challenging to alter, repair and duplicate. 
In recent decades, computer aided modeling and manufacturing for reliefs have been widely developed. 
Relief production has become more efficient and stable with the introduction of computer-driven milling equipment and 3D printing technology. 
Digital relief modeling has also been developed, especially 3D object based modeling technology can generate high-quality relief models. 
\textbf{However, this technology is dependent on the acquisition of high-quality 3D objects, which is not an easy task}.

Despite the advancement of 3D modeling, computer-driven milling equipment, and 3D printing technology, 
the design of relief sculptures is still largely in the hands of artists. 
Generally speaking, the digital relief is typically represented as a height map and the artist can use an image to assist the relief design. 
An artist carefully assigns a set of sparse feature points and construct a set of curves to match the background image in order to create a height map. 
The production heavily depends on the skills and patience of the artist. 
\textbf{Why not liberate our hands and automatically generate the relief shape automatically from an image?}

This work focuses on the problem of human relief generation from a single color photograph. 
This solution significantly reduces the cost of relief modeling and makes it more widely applicable. 
However, this problem is extremely challenging. 
To facilitate relief modeling, this paper presents an automatic and instant solution to create digital relief from a single photograph. 
Specifically, we focus on automatic generation of portrait relief from a full-body human photograph.

Relief generation from a single RGB color photograph is fascinating but challenging. The major challenges are as follows: 

\begin{myitemize}[\textbullet]
\item One of the most difficult challenges in making this task possible is the lack of a photo-relief dataset. 
The relief modeling is an artistic creation, not as easily accessible as photographs. 
As a time-consuming and laborious task, it is not realistic to let artists create enough data manually. 
Moreover, there are considerable differences in the creations due to different aesthetics, which may be not conducive to machine learning. 
Therefore, no ground-truth data can be available for supervised learning.

\item The second difficulty is that there are infinite variations of human portraits in terms of poses, shapes, hair styles, garments, and accessories. 
The important visual cues in the photo need to be recognized and properly presented in a very limited height range, 
such as the facial features, hair, the silhouettes between objects and parts, and the garment wrinkles.

\item The photo is the result of a complex combination of geometry, material, texture, and lighting. 
As the resulting height is greatly squeezed, the geometry in the relief is inconsistent with the geometry of the original 3D human body, 
which increases the difficulty of disentangling geometry and color. 
In summary, this task is to robustly extract geometric structure and details represented by a very limited height range 
from a photograph taken under various complex lighting conditions, indoors or outdoor.
\end{myitemize}

For the general scenes, it is not possible to keep geometry and details fidelity within a compressed range. 
There is a strong tension between the goal of preserving the appearance of the shape and the requirements that the shape be continuous and greatly flattened for effective relief. 
However, it is gratifying that a less accurate shape represented by a greatly compressed height range is still acceptable for the portrait relief design. 

Our network specifically aims to address the above-mentioned challenges. 
In this paper, we present the first attempt to automatically generate a relief shape from a full-body human photograph. 
Due to the lack of ground-truth of relief data, this task is accomplished by training a neural network from depth maps of 3D scenes with a specially designed loss function. 

\noindent \textbf{\\Contributions.}
The main contributions of this paper are summarized as follows:

\begin{myitemize}[\textbullet]
\item Our method generates high quality digital human relief instantly from a single photograph, using a strategy that does not rely on ground truth relief data.
\item Collect a batch of 3D human models with various poses synthesized from scanned 3D human figures, combine them into various scenes, and render them to generate a large-scale dataset under various lighting.
\item The proposed gradient-based losses significantly contribute to extract the geometric structure and fine features (e.g., garment wrinkles) as well.
\end{myitemize}

It is important to note that the goal of this work is not to replace artists, but rather to provide easy and efficient tools for creating human reliefs for general users.

This paper is organized as follows:
Section 2 describes related work.
Next in Section 3 we introduce the architecture of our networks and the implementation details of our methodology.
In Section 4, some experimental results and comparisons are shown.
Finally, we conclude our paper with future work in Section 5.

\section{Related Work}

In this section, we briefly review the relief generation methods that are most related to our work. 
Automatic or semi-automatic relief modeling has been a subject of interest in computer graphics. 
We subdivide the existing works into object-based methods and image-based methods.

\textbf{Relief modeling from 3D objects}.
Object-based methods treat a 3D scene as a height/range field converted from its depth field, and transform the height field into relief directly on height field\cite{CMS97,SPRF09}, in gradient domain \cite{WDBRF07, kbs07, Ker07, KTRAH09, BianH11}, in normal domain \cite{JMS14, WTP18}, and in manifold domain \cite{ZZZY13, ZZLLZ15}.
The key manipulation of most gradient-based methods is to clip and attenuate the gradient magnitudes explicitly or implicitly. 
The normal-based methods compress the height field implicitly and represent the fine details in normal domain, which makes the resulting relief exhibit a very similar appearance with the input 3D object.
The mesh-based methods consider the relief generation in manifold domain, and details manipulation and preservation, height compression, height discontinuities removal are totally performed on an involved 3D mesh.


\begin{figure*}
  \centering
  \hspace{-0.258in}
  \subfigure{
    \includegraphics[scale=0.31]{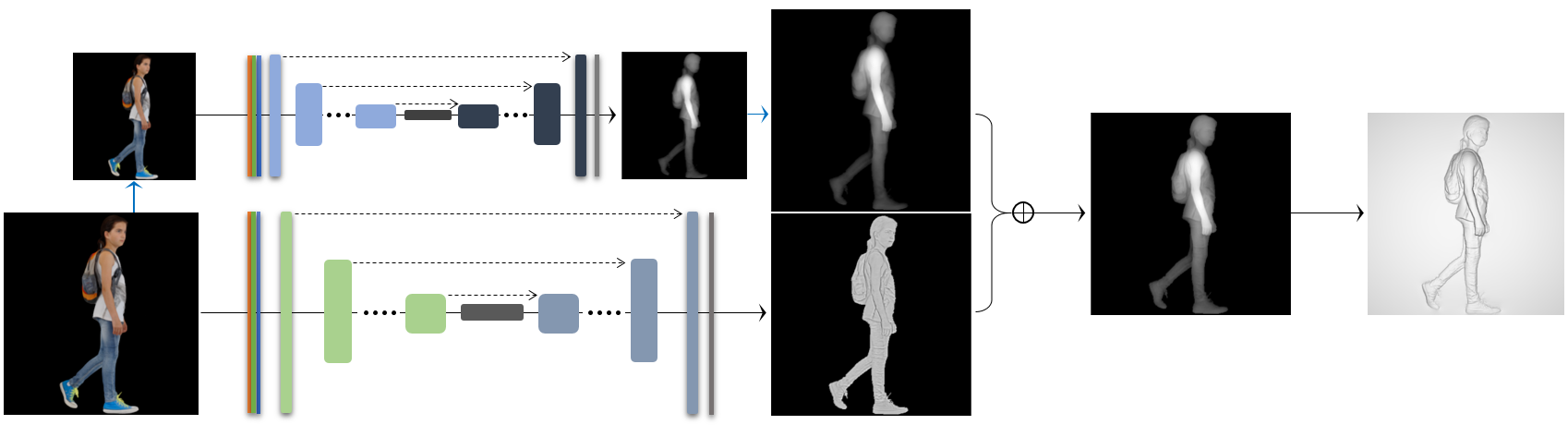}}
  \caption{\label{fig:network}
The architecture of our network consists of a structure path and a detail path. The dashed arrows represent skip connections.
The gray dots indicate repetition of the dense blocks for down-sampling layers and up-sampling layers.
}
\end{figure*}

\textbf{Relief modeling from images}.
Another kind of methods focus on generating reliefs from 2D images. 
These methods are often restricted to generating reliefs from particular images, such as face photos, calligraphy images, etc. 
The techniques of shape-from-shading (SfS) and neural network are used to develop a method for creating relief from a single human face image \cite{wmr13}.
A mesh-based modeling method is proposed to generate Chinese calligraphy reliefs from a single image \cite{ZCLJZ18}. 
The relief is constructed by combining a homogeneous height field and an inhomogeneous height field together via a nonlinear compression function. 
A template-based method for portrait relief modeling from a single image is proposed \cite{ZZWCJH2019}. 
Given a portrait image, a template face is first used to fit the portrait, then bi-Laplacian mesh deformation is applied to align the facial features, and SFS-based reconstruction with a few user interactions is finally used to optimize the face depth, and create a relief with similar appearance to the input.
Recently, a complex framework is presented for producing human relief from a single photograph \cite{YCZCZ21}.
Given an input photo of one or multiple persons, 3D skeletons and 3D guide models are detected to reconstruct an overall base shape. 
And a fine-scale normal map integrated with the base shape to produce the final relief model. 
Some user interactions need to be introduced to handle the occlusion relationship and the non-rigidly contour register.
This method has multiple stages and requires some crucial interactions.
Textured tactile relief is another interesting topic. A method which converts the high resolution image of the painting into textured tactile relief is proposed \cite{RMP11}. 
The result is a height map of the topmost surface of the layered depth diagram, which is further sliced into layers of constant thickness that can be assembled on top of each other. 
Similarly, a systematic method is presented to generate 2.5D tactile models \cite{FGVPVC14}. 
This method combines perspective geometry-based scene reconstruction and SFS-based volume reconstruction methods.

\section{Methodology}

In essence, we use a deep neural network
to approximate a function $H_r = \Phi (I)$ that maps a single photograph to a height field representing relief.
In this section, we describe the design methodology of the proposed network, including the network architecture, dataset construction, loss function, and details of implementation as well as network training.

\subsection{Network Construction}

3D objects can be decoupled into overall shapes and local details, so the conversion from 3D objects to reliefs can also be decomposed into two parts. 
In this paper, these two parts are called structure layer and detail layer respectively.

To make a clear division of labor in network modules, a two-scale architecture is proposed to create high-quality relief from a single photograph. 
The most important perceptual features that are retained throughout the conversion of a 3D object to its relief are kept in the structure layer, 
including shading under incident illumination and silhouettes at depth discontinuities. 
The structure layer conveys the central visual cues of the global shape while ignoring the fine details.
The detail layer focuses more on the local grainy details. 
To this end, we introduce a filter similar to difference of Gaussians (DoG).

For a general photograph with a resolution of about $1024$, structure network can generate a reasonable relief shape with appropriate details. 
Detail network is optional, but when the resolution of the input photograph is larger than $1024$, it can be used to introduce more grainy details. 
The detail network focuses on adding more subtle details at a fine level, which is useful for input photos with a resolution larger than $1024$. 
Given a large photograph (such as 2k resolution), the structure network takes as input a downsampled $1024 \times 1024$ image, and produces a structure layer of $1024 \times 1024$ resolution. 
The detail network takes as input the original image, and produces a detail layer. 
Then the structure layer and detail layer are fused into a relief image with the original resolution of input photo. 

In this paper, we utilize a DenseNet-like architecture \cite{HuangLMW17} to aggregate features
of different scales to extract fine details while removing height gaps in an efficient way.
The architecture of our network is illustrated in Fig. \ref{fig:network}.

\subsection{Dataset Construction}

Unlike other tasks in computer vision, there are no `ground truth' data for relief data.
Therefore, the first step of our method is to construct a dataset containing enough photo-heights.
To this end, we render a batch of 3D human models and take them as the `ground truth' data.
Specifically, our synthetic human photo-height dataset consists of many pairs of rendering images and heights of 3D human models. 

\begin{figure}
  \centering
  \hspace{-0.15in}
  \subfigure{
    \includegraphics[scale=1.005]{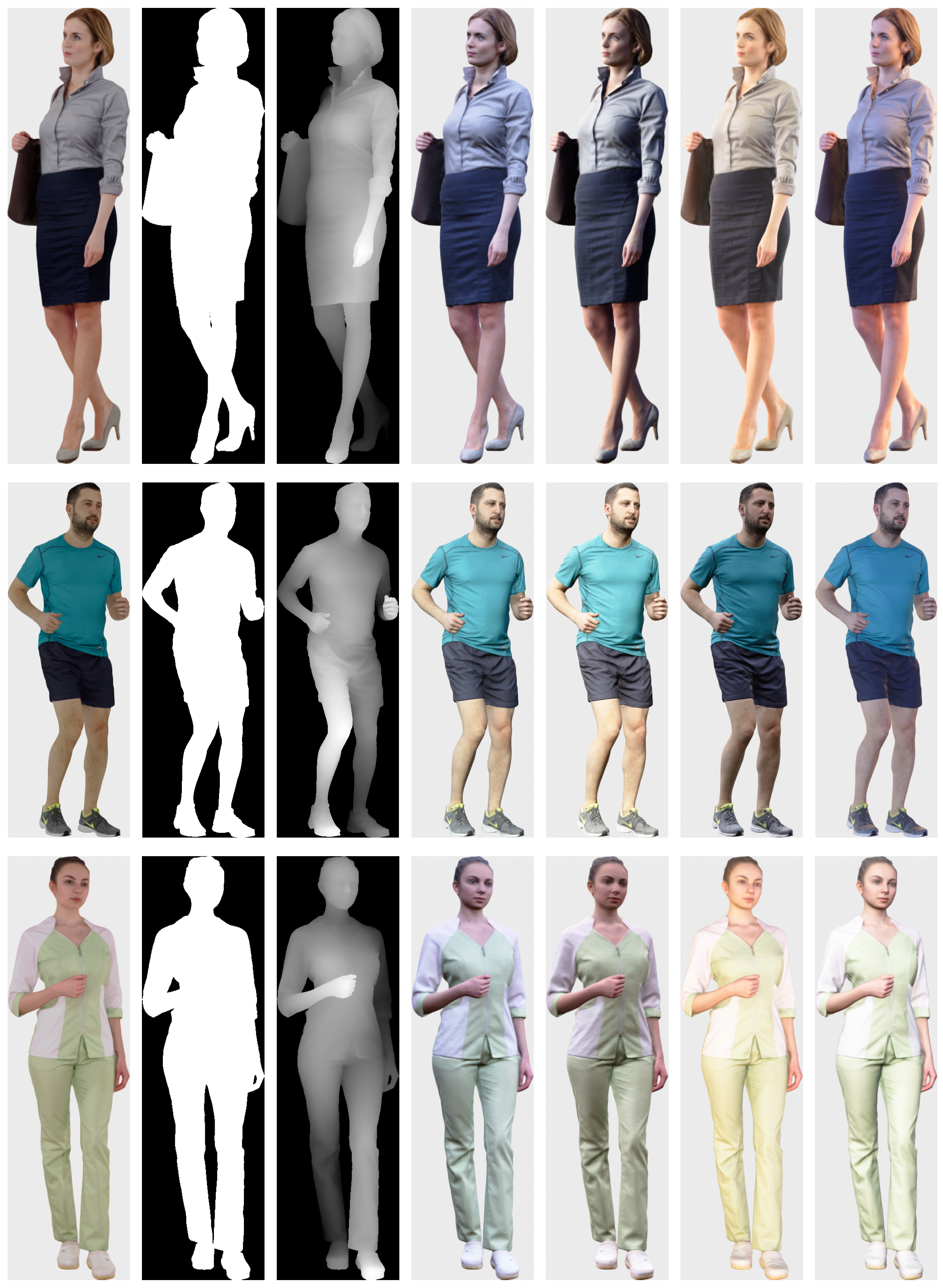}}
  \caption{\label{fig:dataset}
Examples from our human photo-height dataset. For each human figure, an albedo map, binary mask, height map, and several rendering maps with different illuminations are displayed.
}
\end{figure}

We choose some scanned and synthesized 3D human models and then construct 500 simple 3D scenes which contain one to three persons.
Given a 3D scene, we can capture a batch of rendering images and height fields from various viewpoints.
We capture each 3D scene along 20 directions.
In order to imitate the photographs taken `in the wild', we use image-based lighting technology to render the 3D human scenes in our dataset. 
We used the synthetic rendering images of $1024 \times 1024$ pixels for training our SNet. 
Specifically, from the 20 illuminations of different outdoor locations and weather conditions, we randomly pick several illuminations for rendering each 3D human scene.
Finally, we obtain about 50000 pairs of rendering images and heights as training set. 
Three examples rendered under 4 different illuminations are displayed in Fig. \ref{fig:dataset}.


\subsection{Loss function}

Given a color photograph, our network aims at mapping RGB colors to relief height values, 
which makes the characters in the photograph stand up slightly.
However, the character reliefs corresponding to photographs are very rare in reality.
Therefore, we use 3D textured human models to learn constructing character reliefs in an unsupervised manner.

Therefore, our method must collapse empty space at object silhouettes, essentially
squashing foreground objects down against the background objects that they occlude. 
Fortunately, this requirement supports the goal of reducing dynamic range.
This compression takes place in the gradient domain and is followed by an integration step
that recovers a height field of the resulting relief from the modified gradients.

\textbf{Gradient manipulation}

Due to lack of ground truths, the loss functions defined in the height domain,
e.g., Mean Squared Error (MSE) between true and predicted height values, are not available.
To avoid explicitly constructing the pseudo ground truths, 
our method can optimize the height field in gradient domain and normal domain.

Our method begins with a textured 3D scene as input, renders it into a color picture,
and samples it into a raw height field at a specific resolution. 
This height field is not our goal, but we can take it to assist in training the network.
Specifically, we train the network by optimizing the output height field to meet the requirements of relief form in the gradient domain. 

Given a depth map sampled from a 3D scene, Weyrich et al. \cite{WDBRF07}
compress its gradient magnitudes nonlinearly and reconstruct the relief height field by solving a Poisson equation. 
To remove height discontinuities, they introduce a threshold to detect and remove the silhouettes, 
and then they apply a nonlinear function to compress the remaining gradient magnitudes. 

We do not aim to compute reliefs for supervised learning, 
\textbf{one reason is that it is computationally expensive to solve large systems of linear equations for all data in the training set}. 
We implemented the algorithm with Intel Math Kernel Library (MKL) on a PC with an Intel(R) Core(TM) i7-7800X @ 3.5 GHz, 32 GB of RAM. 
For a height map with a resolution of $1024 \times 1024$, the running time of relief generation algorithm is about $16$ seconds. 
Another reason is that we define loss functions in the gradient domain, so the relief heights are not required. 
\textbf{We focus on manipulating all gradients tactfully without introducing thresholds to remove silhouettes explicitly}, and using the modified gradients to train our neural networks. 
A number of experiments have revealed that employing the following sigmoid variant function works well for our task. 

$$
\mathcal{S}(x,\alpha) = \frac{1-e^{-\alpha x}}{1+e^{-\alpha x}}
$$

\begin{figure}
  \centering
  \hspace{-0.258in}
  \subfigure{
    \includegraphics[scale=0.43]{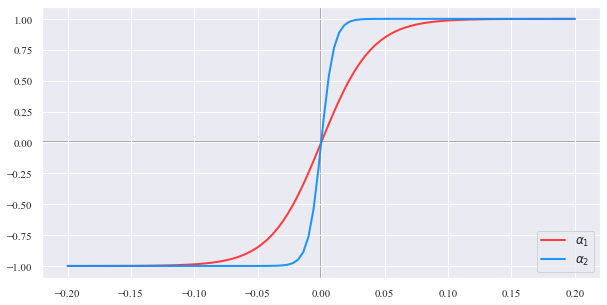}}
  \caption{\label{fig:loss_sig}
Two variants of the sigmoid function compared. The function pushes more values to $1$ as the parameter $\alpha$ increases.
}
\end{figure}

By simple substitution, it can be obtained that $(\mathcal{S}(x,1)+1)/2$ is equal to the sigmoid function. 
An appropriate parameter $\alpha$ can boost small gradients while attenuating large slopes. 
Two graphs of this family of functions are shown in Fig. \ref{fig:loss_sig}.


The structure of relief is presented by the remaining interval in the height field.
These intervals together form the level of height in the entire relief shape, and the remaining features show important visual cues.

\subsection{Structure network (SNet)}

Due to the lack of ground-truth of relief data, we present a stratege to generate high quality digital human relief from a single photograph by utilizing the height fields. 
Although the original height fields of 3D human objects cannot be used directly for supervised learning, we can utilize them in light of our objectives. As the primary goals are to remove the unexpected discontinuities of a given height field while preserving the fine details, we accomplish them by using loss functions in the gradient domain. Specifically, for a set of inputs $\{\mathbf{I}\}$ and their height fields $\{\mathbf{h}_{i}\}$, we learn the network parameters $\Theta$ via regression:

\begin{equation}{\label{eqn:lossfunction1}}
\mathcal{L}^{s} = \frac{1}{N}\sum_{(u, v) \in \Omega} \mathcal{E} \left(\nabla \mathbf{h}_{s}(u, v),\nabla \mathbf{h}_{i}(u, v)\right) ,
\end{equation}

where $\Omega$ indicates the definition domain of the relief in $\mathbb{R}^2$, $\mathbf{h}_{s}(u,v) = \mathcal{N}_{s}(\mathbf{I}(u,v); {\Theta}_{s})$ is the predicted height of stucture surface at $(u,v)$, and $\mathcal{E}$ is used to estimate the error between the two gradient vectors $\nabla \mathbf{h}_{s}(u, v)$ and $\nabla \mathbf{h}_{i}(u, v)$. We are not calculating the difference between them directly. Actually, we first manipulate the gradient vector $\nabla \mathbf{h}_{i}(u, v)$ and then estimate the errors in different ways.

We found that $L_1$ norm and $L_2$ norm resulted in significantly different effects, so we used them to train models separately. The specific loss functions are defined as follows,

\begin{equation}{\label{eqn:lossfunction2}}
\mathcal{L}_{1}^{s} = \frac{1}{N}\sum_{(u, v) \in \Omega}\left\|\nabla \mathbf{h}_{s}(u, v)-\mathbf{g}_{i}(u, v)\right\|_{1},
\end{equation}

and

\begin{equation}{\label{eqn:lossfunction3}}
\mathcal{L}_{2}^{s} = \frac{1}{N}\sum_{(u, v) \in \Omega}\left\|\nabla \mathbf{h}_{s}(u, v)-\mathbf{g}_{i}(u, v)\right\|_{2}^{2},
\end{equation}

where $\mathbf{g}_{i}(u, v) = \Phi_{1}\left(\nabla \mathbf{h}_{i}(u, v), \alpha\right)$, and $\Phi_1$ manipulates the gradient magnitudes using the sigmoid variant function defined above,

$$
\Phi_{1}(\mathbf{x}, \alpha) = \mathcal{S}(\|\mathbf{x}\|, \alpha) \mathbf{x}.
$$

It is worth noting that our network specifically targets the modified gradients $\Phi_{1}\left(\nabla \mathbf{h}_{i}(u, v), \alpha\right)$, rather than the original height field $\mathbf{h}_{i}(u,v)$.

\textbf{Cosine loss function}

Additionally, we also experimented with the loss function defined by the cosine distance between the output normal and the target one,

\begin{equation}{\label{eqn:lossfunction3}}
\mathcal{L}^{n} = 1 - \frac{1}{N}\sum_{(u, v) \in \Omega} \langle {\bf n}_{s}(u, v), {\bf n}_{i}(u, v) \rangle,
\end{equation}

where the normal vector ${\bf n}_{i}(u, v)$ is derived from the modified gradients $\mathbf{g}_{i}(u, v)$,

$$
{\bf n}_{i}(u, v) = \dfrac{\langle -\mathbf{g}_{i}(u, v), \eta \rangle}{\| \langle -\mathbf{g}_{i}(u, v), \eta \rangle \|},
$$

where $\eta$ is a constant that controls the steepness of normal vectors.

In Section 4, we will compare the specific results generated by employing these various loss functions.

\subsection{Detail network (DNet)}

The goal of the above structure network is to learn large visual cues which convey shape. 
In order to compensate for the grainy details, we introduce a detail network which is trained with the following loss function,

\begin{equation}{\label{eqn:lossfunction2}}
\mathcal{L}^{d} = \frac{1}{N}\sum_{(u, v) \in \Omega}\left\| \mathbf{h}_{d}(u, v)- \text{div}\left[ \Phi_{2}\left( \nabla \mathbf{h}_{i}(u, v), \alpha_1, \alpha_2\right)\right]\right\|_{2}^{2},
\end{equation}

where $\mathbf{h}_{d}(u,v) = \mathcal{N}_{d}(\mathbf{I}(u,v); {\Theta}_{d})$ is the predicted details, and $\Phi_2$ is used to extract the detail gradient magnitudes from $\mathbf{h}_{i}$ in a DoG-like band-pass manner,

$$
\Phi_{2}(\mathbf{x}, \alpha_1, \alpha_2) = \left(\mathcal{S}(\|\mathbf{x}\|, \alpha_2) - \mathcal{S}(\|\mathbf{x}\|, \alpha_1)\right) \mathbf{x}.
$$

Subtracting one gradient field from the other extracts gradients whose magnitudes lie in a certain range while attenuating or even eliminating the gradients that are far from the band center.


\begin{figure}
  \centering
  \hspace{-0.1in}
  \subfigure{
    \includegraphics[scale=1.0]{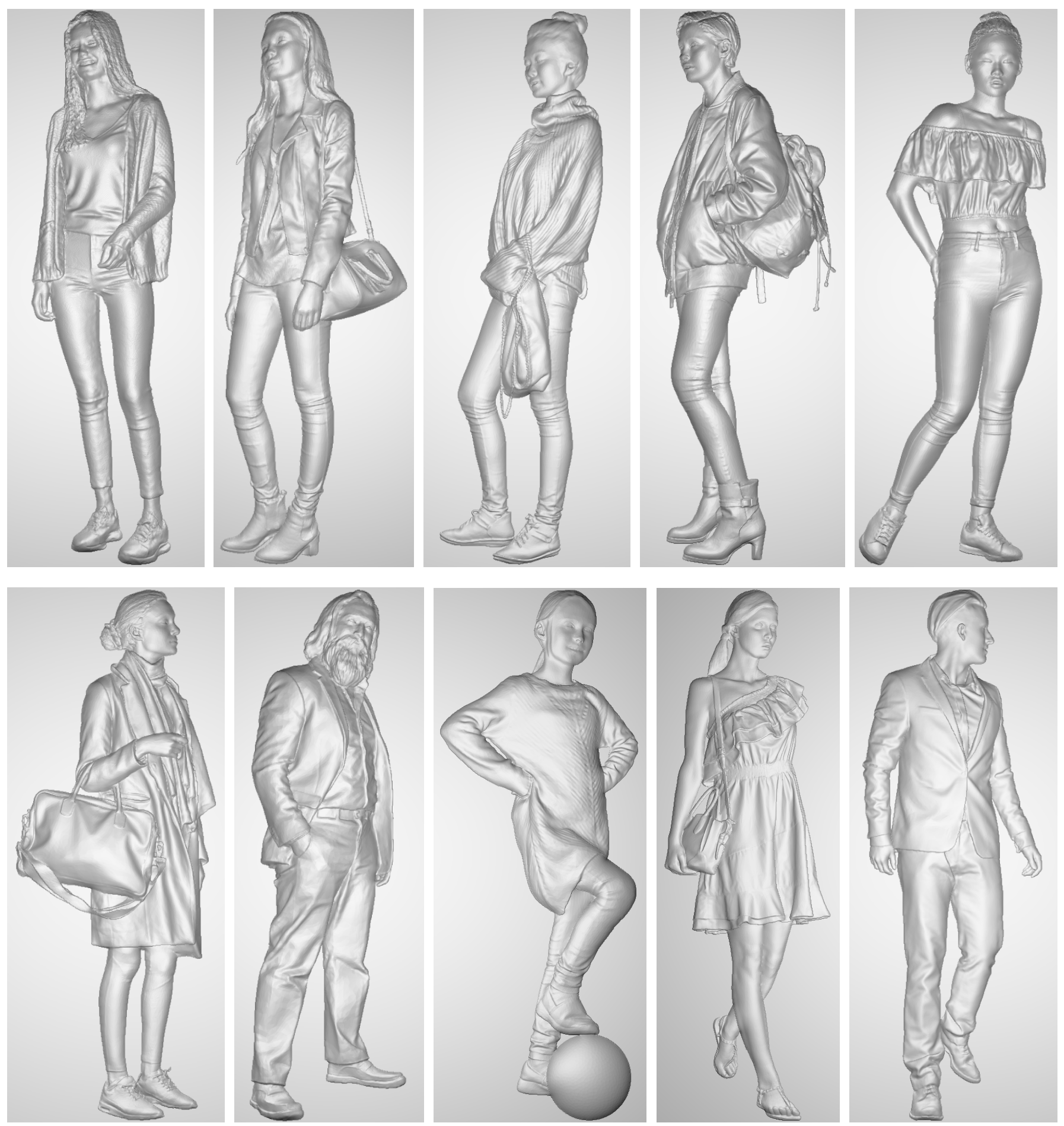}}
  \caption{\label{fig:recexamples}
Some visual examples of results reconstructed from the modified gradients.
}
\end{figure}

\section{Results}

Our networks are trained from scratch and a validation set is used to avoid overfitting by early stoping the training. 
The networks are implemented in PyTorch with a GeForce GTX 3090 Ti GPU. 
We first train the network on the $\mathcal{L}_{2}^{s}$ loss for 30 epochs using ADAM optimizer with a learning rate of $2\mathrm{e}{-4}$, then we fine-tune the trained network on $\mathcal{L}_{1}^{s}$ loss and $\mathcal{L}^{n}$ loss for 5 epochs respectively. 

In this section, we present prediction results for relief reconstruction on some synthetic test data, 
as well as the results for real photographs.
At inference stage, we take as input a photograph (RGB image) of one or more persons and the corresponding binary mask image of persons. 

\begin{figure*}
  \centering
  \hspace{-0.1in}
  \subfigure{
    \includegraphics[scale=1.0]{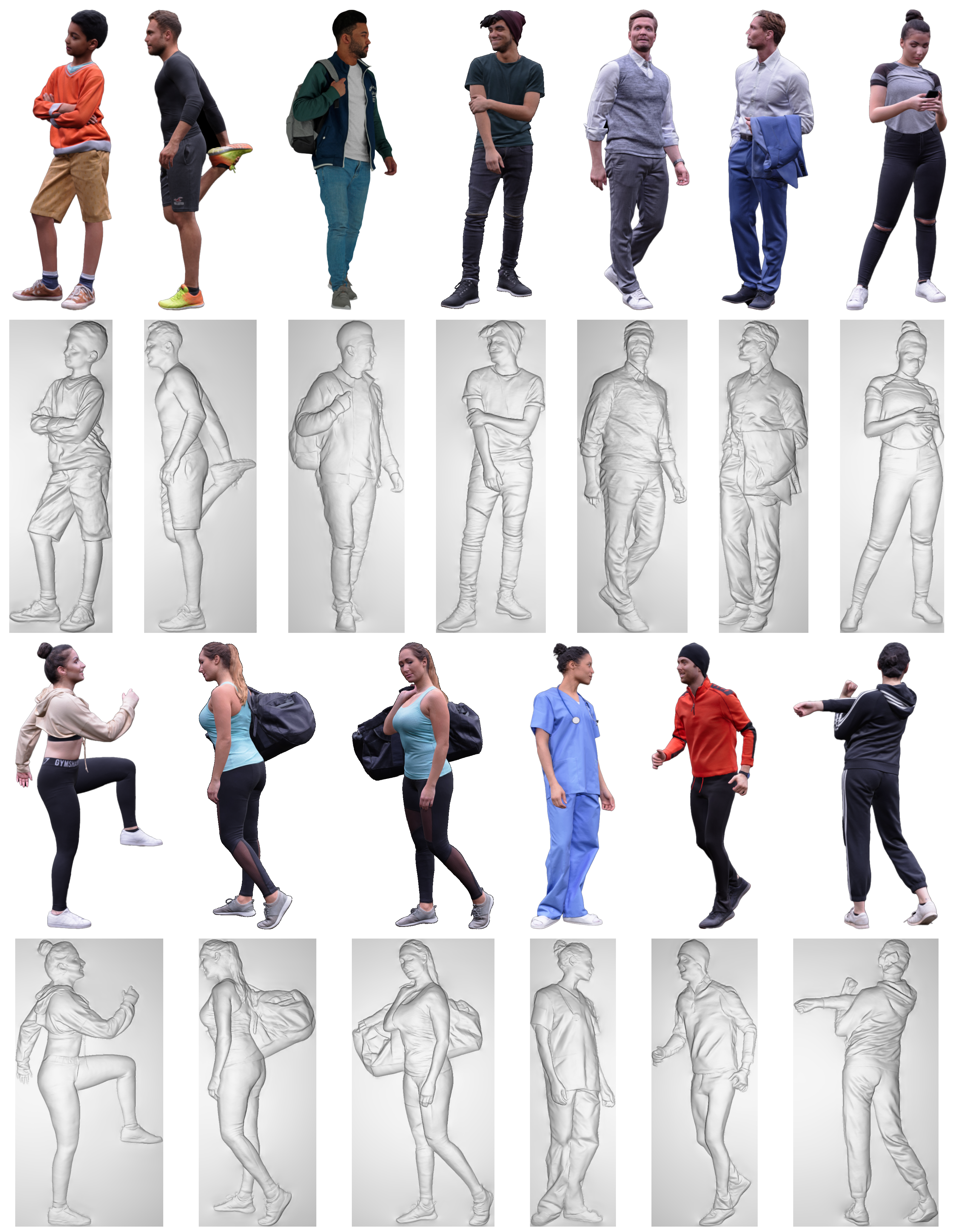}}
  \caption{\label{fig:synthetic}
More examples generated from synthetic data.
}
\end{figure*}

First, we present the prediction results on some synthetic test images. 
These synthetic images are rendering images of 3D human models by using image-based lighting technology. 
Fig. \ref{fig:synthetic} gives some results produced by our SNet trained with the $\mathcal{L}_{2}^{s}$ loss.
The structure layer keeps the most relevant perceptual features which are preserved during the conversion from a color image to its height map, 
such as silhouettes at depth-discontinuities and shading under incident illumination.

\begin{figure}
  \centering
  \hspace{-0.1in}
  \subfigure{
    \includegraphics[scale=0.2750]{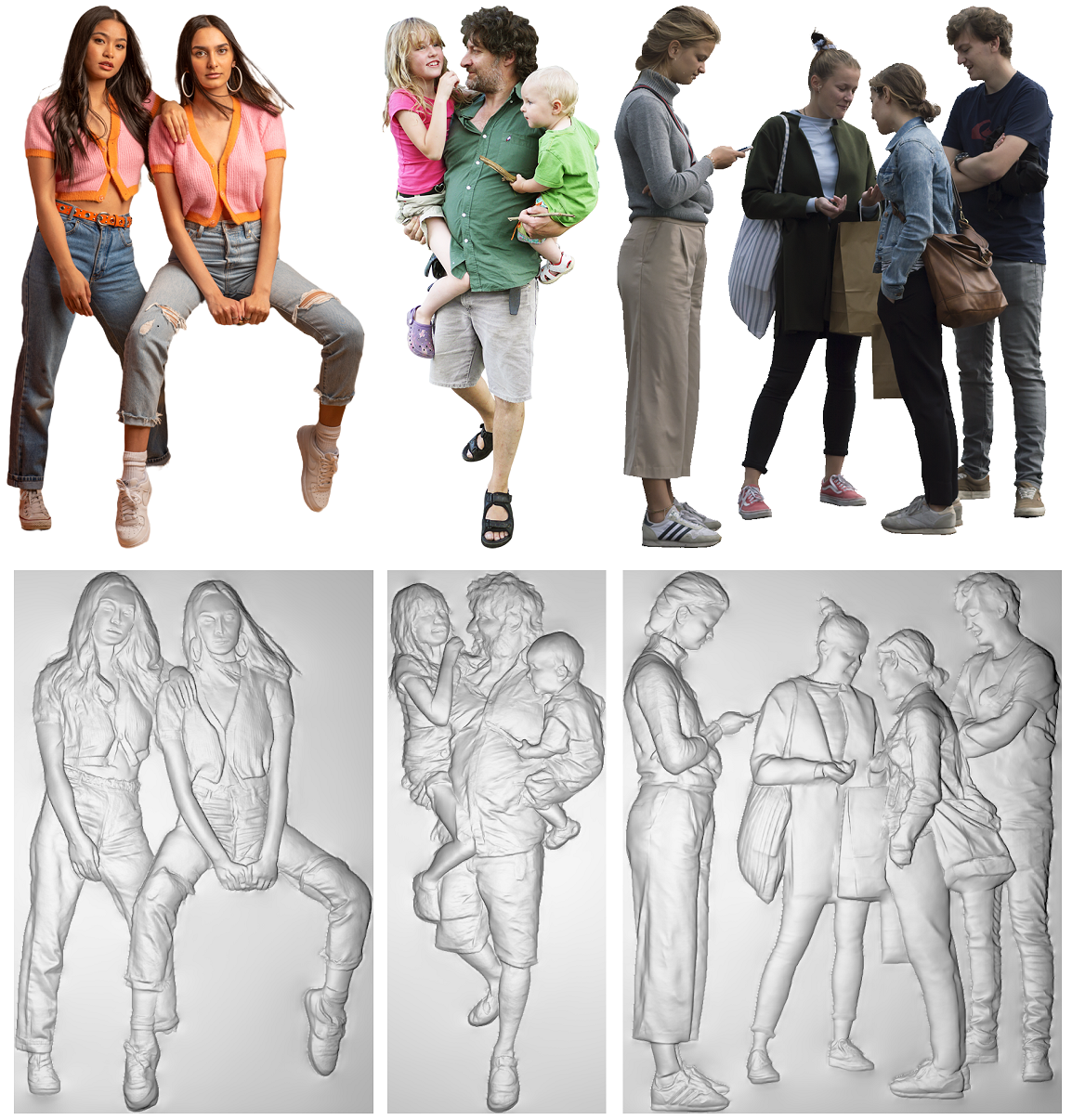}}
  \caption{\label{fig:multi}
Predictions on real photographs of multiple persons.
}
\end{figure}

\begin{figure*}
  \centering
  \hspace{-0.1in}
  \subfigure{
    \includegraphics[scale=1.0]{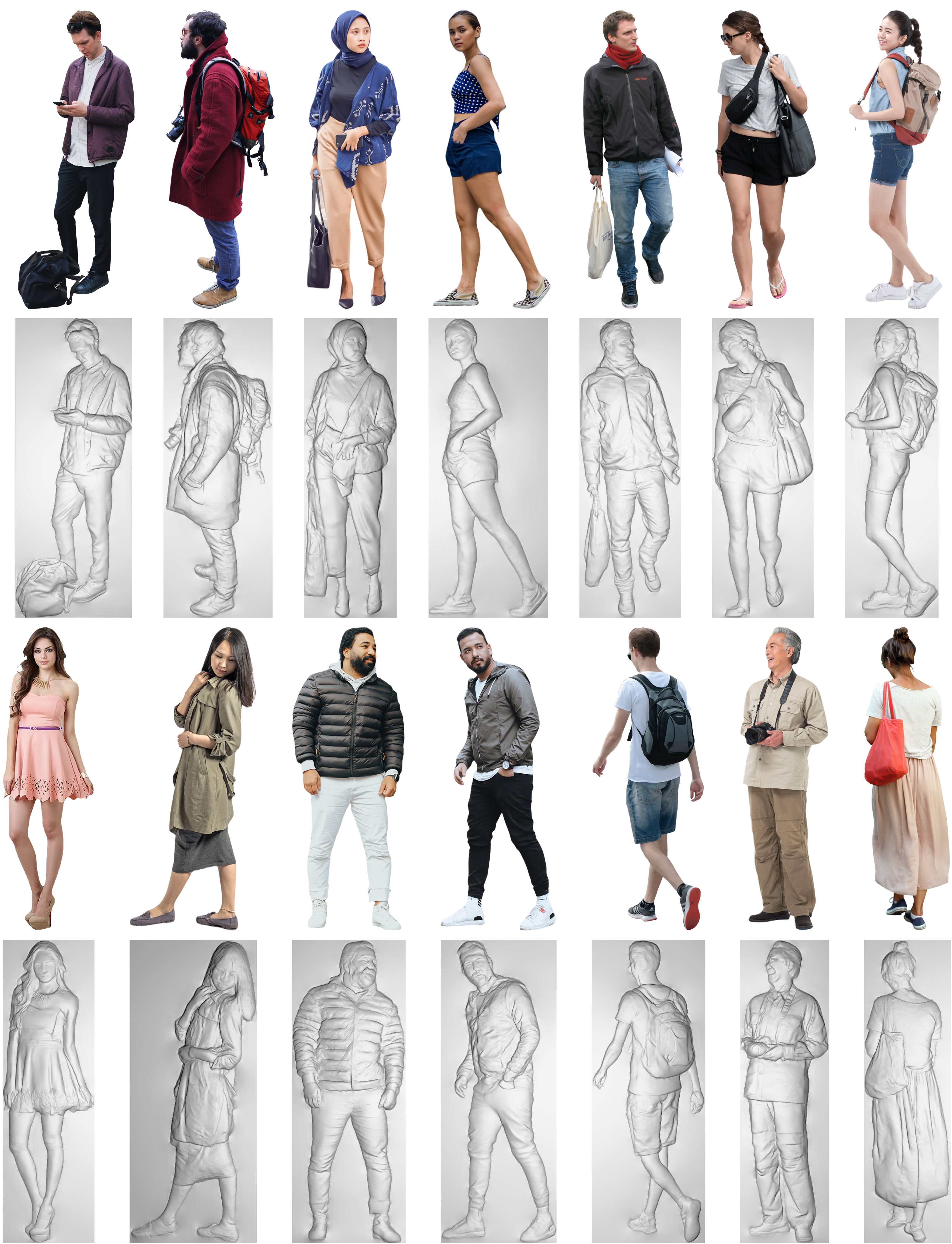}}
  \caption{\label{fig:realphoto}
More examples generated from real photographs.
}
\end{figure*}

In Fig. \ref{fig:realphoto}, we present the reliefs predicted from some real photographs.
Our model can also process complex scenes containing multiple persons.
Fig. \ref{fig:multi} gives three real photographs and the predicted reliefs. \
There are 2, 3 and 4 persons in these photos respectively.

\begin{figure}
  \centering
  \subfigure[Input]{
    \label{fig:loss:a}
    \includegraphics[angle=0,width=0.6704in]{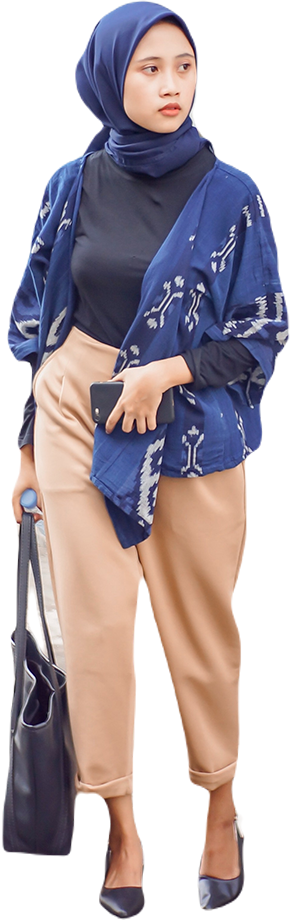}}
    \hspace{-0.123in}
  \subfigure[$\mathcal{L}_{2}^{s}$]{
    \label{fig:loss:b}
    \includegraphics[angle=0,width=0.6726in]{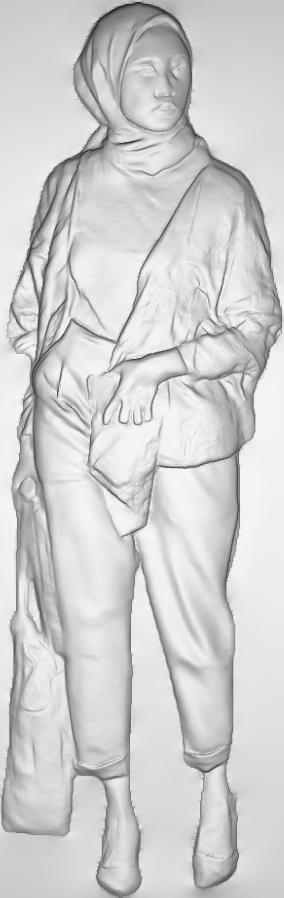}}
  \subfigure[$\mathcal{L}^{n}$]{
    \label{fig:loss:c}
    \includegraphics[angle=0,width=0.6726in]{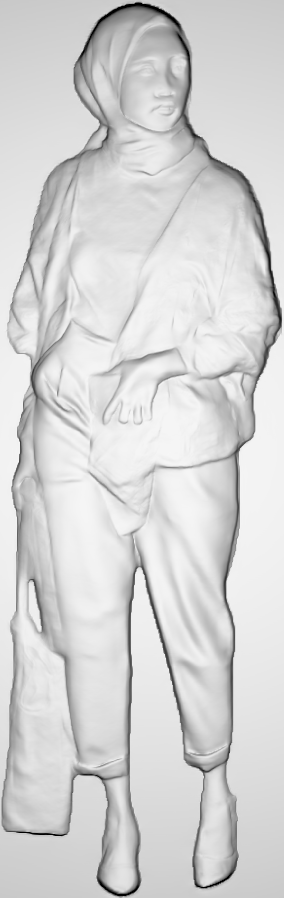}}
    \hspace{-0.119in}
  \subfigure[]{
    \label{fig:loss:d}
    \includegraphics[angle=0,width=1.2843in]{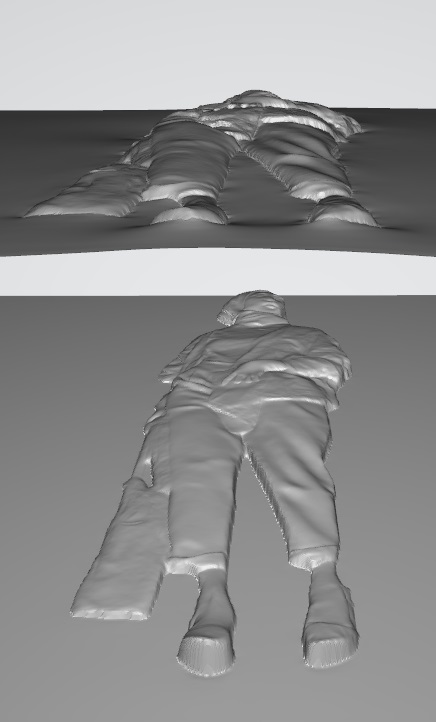}}
  \caption{\label{fig:loss}
Comparisons of different loss functions.
}
\end{figure}

\textbf{Different loss functions.}
To generate different visual appearances of the resulting reliefs, we introduce three loss functions, including $\mathcal{L}_{1}^{s}$, $\mathcal{L}_{2}^{s}$ and $\mathcal{L}^{n}$. The networks trained on these losses can generate reliefs with significantly different visual appearances. Because there is no ground truth for relief, it is actually difficult to evaluate the qualities of resulting reliefs. Due to the subjectivity of aesthetics, we tend to present different visual appearances as different visual styles for users to choose freely. 
As shown in Fig. \ref{fig:loss}, the loss $\mathcal{L}_{2}^{s}$ allows non-zero background heights along the silhouettes which is more pronounced under illumination. On the contrary, a perfectly flat background can be enforced by the losses $\mathcal{L}_{1}^{s}$ and $\mathcal{L}^{n}$ without using explicit constraints at silhouettes between scene elements and the back plane. 
Small fluctuations in the background plane may cause smaller errors in the gradient domain, but can introduce larger cosine errors in the normal domain, so the loss $\mathcal{L}^{n}$ results in a flatter background. Furthermore, an abrupt foreground is formed at silhouettes to fit the normals of the foreground. 

\begin{figure}
  \centering
  \hspace{-0.099in}
  \subfigure[]{
    \label{fig:2k:a} 
    \includegraphics[angle=0,width=0.8in]{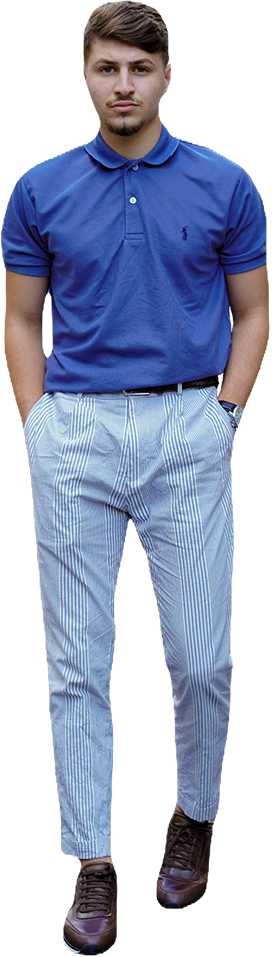}}
  \hspace{-0.056in}
  \subfigure[]{
    \label{fig:2k:b} 
    \includegraphics[angle=0,width=0.808in]{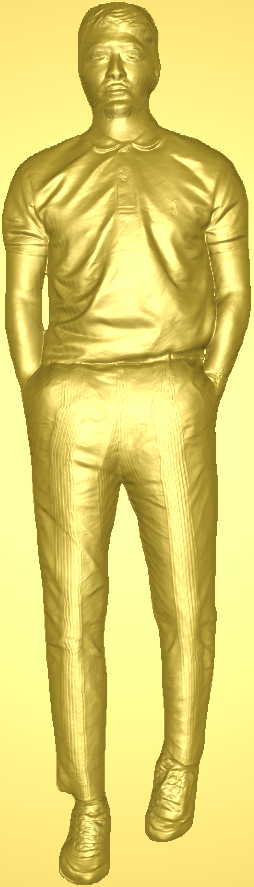}}
     \hspace{-0.056in}
  \subfigure[]{
    \label{fig:2k:c} 
    \includegraphics[angle=0,width=0.801in]{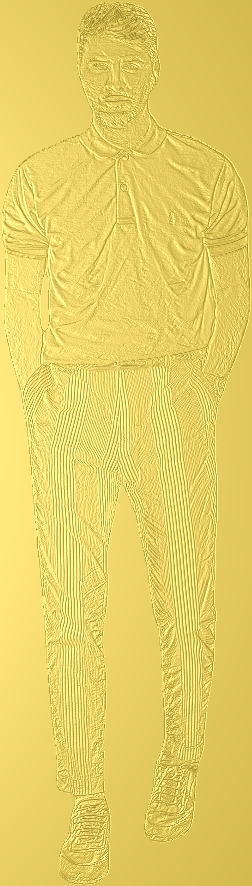}}
    \hspace{-0.056in}
  \subfigure[]{
    \label{fig:2k:d} 
    \includegraphics[angle=0,width=0.7986in]{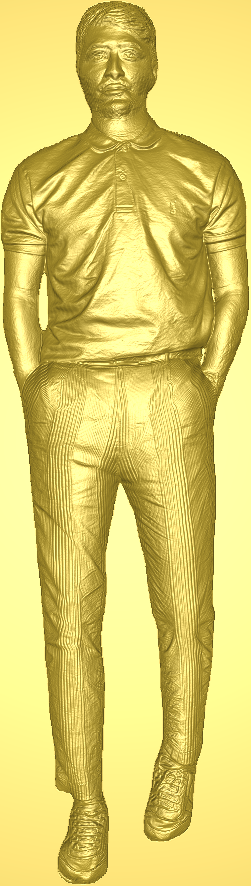}}
  \caption{\label{fig:2k}
Relief generation from a photograph with 2k resolution. 
(a) An 2048 resolution photo;
(b) relief generated using SNet from a 1024 resolution image downsampled from (a);
(c) detail generated using DNet from (a);
(d) the resulting relief of fusing (c) with an upsampled version of (b).
}
\end{figure}

\textbf{Relief from high-resolution photographs.} Taking the GPU memory and computation time into account, we use $1024 \! \times \! 1024$ resolution images for training our models.
For an image with 2K or higher resolutions, we use a simple strategies to reconstruct the layers of structure and detail. 
We first downsample it to 1024 resolution, then resample the prediction of SNet to the original resolution, and finally fuse it with the prediction of DNet. 
Experiments show that such a simple scheme can yield high-quality results.
Fig. \ref{fig:2k} shows the resulting relief generated using our models from an input image with 2K resolution.
From left to right, they are photograph, structure layer, detail layer, and the fused relief respectively.

\subsection{Comparison}

To our knowledge, this is the first work of end-to-end trainable neural network for reliefs generating from a single photographs.
Experiment results and comparisons to previous methods will be shown in this section.

Recently, a semi-automatic method is presented for producing human relief from a single photograph \cite{YCZCZ21}.
Given an input photo of one or multiple persons, their method first estimates a 3D skeleton and generates a 3D guide model for each person. 
It warps the projected contours of the guide model with the detected body contours in the image. 
Then the normals of the 3D guide model are warped to reconstruct an overall base shape. 
Finally, a fine-scale normal map integrated with the base shape to produce the final relief model. 
Some user interactions need to be introduced to handle the occlusion relationship and the non-rigidly contour register.
In short, it is a complex system with multiple stages and crucial interactions.
Instead, our method can convert photographs directly into relief models through end-to-end networks.
In addition, for our method, no interaction is required.

\begin{figure}
  \centering
  \hspace{-0.1in}
  \subfigure{
    \includegraphics[scale=0.305]{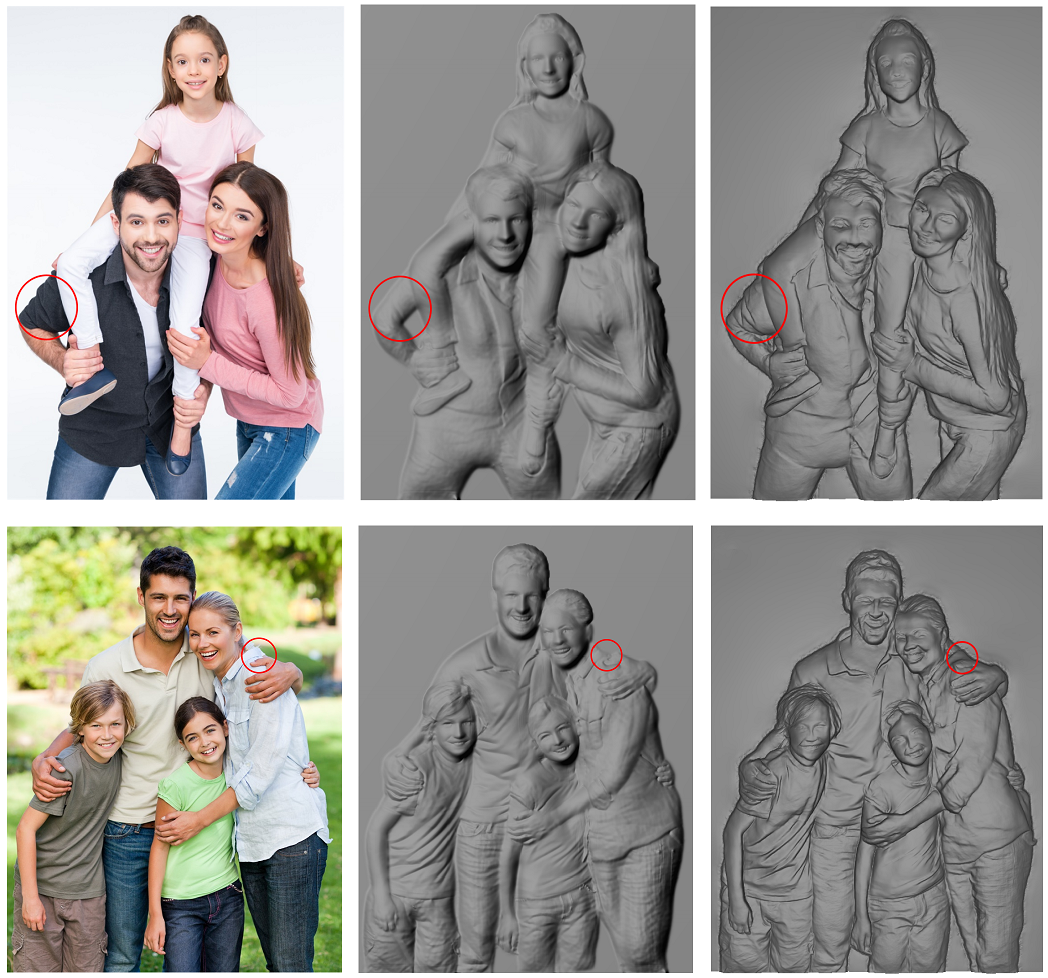}}
  \caption{\label{fig:comparison}
Comparison of the resulting reliefs between our method and the method \cite{YCZCZ21}.
}
\end{figure}

Now, we compare the results of the two methods visually.
As shown in Fig. \ref{fig:comparison}, our results have comparable appearance to their work. 
The results of our method match exactly the pixels of the input image due to the end-to-end learning. 
In their method, contour distortion may occur in some regions due to the intermediate contour matching step (see the regions marked by red circles).
In addition, for these two examples, the total computation time of their method is about several minutes, without taking account of the user interactions. And our method is instantaneous in its prediction.

\section{Conclusions and Future Work}

In this paper, we propose a framework that generates digital relief from a single photograph based on an end-to-end trained deep neural networks. 
A full body human photo-height dataset with diverse poses of humans and various lighting conditions is first constructed, 
image-based rendering technique is used to acquire rendering images under different lighting conditions, 
and the reliefs are constructed from depth maps of scenes composed of 3D human models. 
A two-scale architecture and gradient-based losses are proposed to extract the geometric structure and fine features from a single photograph. 
In current framework, we have focused on photograph of human without background, 
thus relief generation from photographs with various backgrounds will be one of the future work. 
In addition, photographs with  shadows and highlights need to be further investigated in the future.


\bibliographystyle{IEEEtran}
\bibliography{relief}

\end{document}